\documentclass{article}

\usepackage[preprint]{corl_2026} 

\usepackage{booktabs}
\usepackage{multirow}
\usepackage{graphicx}

\usepackage{siunitx}
\usepackage{tabularx}
\usepackage{enumitem}
\usepackage{wrapfig}
\usepackage{amssymb}

\usepackage[capitalize]{cleveref}

\newcommand{\method}{\textsc{Patch Policy}}
\newcommand{\std}[1]{{\scriptsize$\pm #1$}}

\newcommand{\superscript}[1]{\textsuperscript{\textnormal{#1}}}

\title{Patch Policy: Efficient Embodied Control \\ via Dense Visual Representations}

\author{
    \textbf{Gaoyue Zhou\superscript{1}}\thanks{Equal contribution.\quad Corresponding author: \texttt{gz2123@nyu.edu}} \quad
    \textbf{Zichen Jeff Cui\superscript{1}}\footnotemark[1] \quad
    \textbf{Ada Langford\superscript{1}} \quad
    \textbf{Bowen Tan\superscript{1}} \\[2pt]
    \textbf{Yann LeCun\superscript{1,3}} \quad
    \textbf{Lerrel Pinto\superscript{1,2}}
    \\[6pt]
    \superscript{1}Courant Institute, New York University \quad
    \superscript{2}Meta-FAIR \quad
    \superscript{3}AMI Labs
}

\begin{document}
\maketitle


\begin{abstract}
    Pretrained dense visual features from Vision Transformers (ViTs) are powerful yet have been underutilized in robot learning. Modern robot policies either compress each observation into a single global token, or rely on visual backbones trained from scratch, sacrificing both fine-grained spatial detail and the benefits of large-scale visual pre-training. While there exist policies that do operate on dense patch features like large vision-language-action models (VLAs), they tend to be heavy and slow, inheriting the full cost of a billion-parameter vision-language model (VLM) backbone. We close this gap with \textsc{Patch Policy}, a minimal architectural extension that enables transformer-based policies to consume dense pre-trained patch tokens directly without the computational overhead of a full VLM. At its core is a block-causal attention mask that preserves the temporal causality of standard policies while letting the model attend over many patch tokens per observation, alongside other state information. \textsc{Patch Policy} is lightweight, fast, and highly effective. Across four simulated and three real-world environment suites, our method achieves a 40\% relative improvement over policies using state-of-the-art global-pooled representations. Furthermore, it surpasses fine-tuned OpenVLA-OFT by 18\% while using roughly 0.7\% of the parameters. We believe \textsc{Patch Policy} provides a pipeline for the robotics community to readily leverage continuing progress in visual representation learning, without sacrificing the training efficiency or inference speed required for high-frequency, reactive control. Videos can be viewed at \url{https://patch-policy.github.io}.
\end{abstract}

\section{Introduction}
\label{sec:intro}

\begin{figure*}[t]
\center
\includegraphics[width=\textwidth]{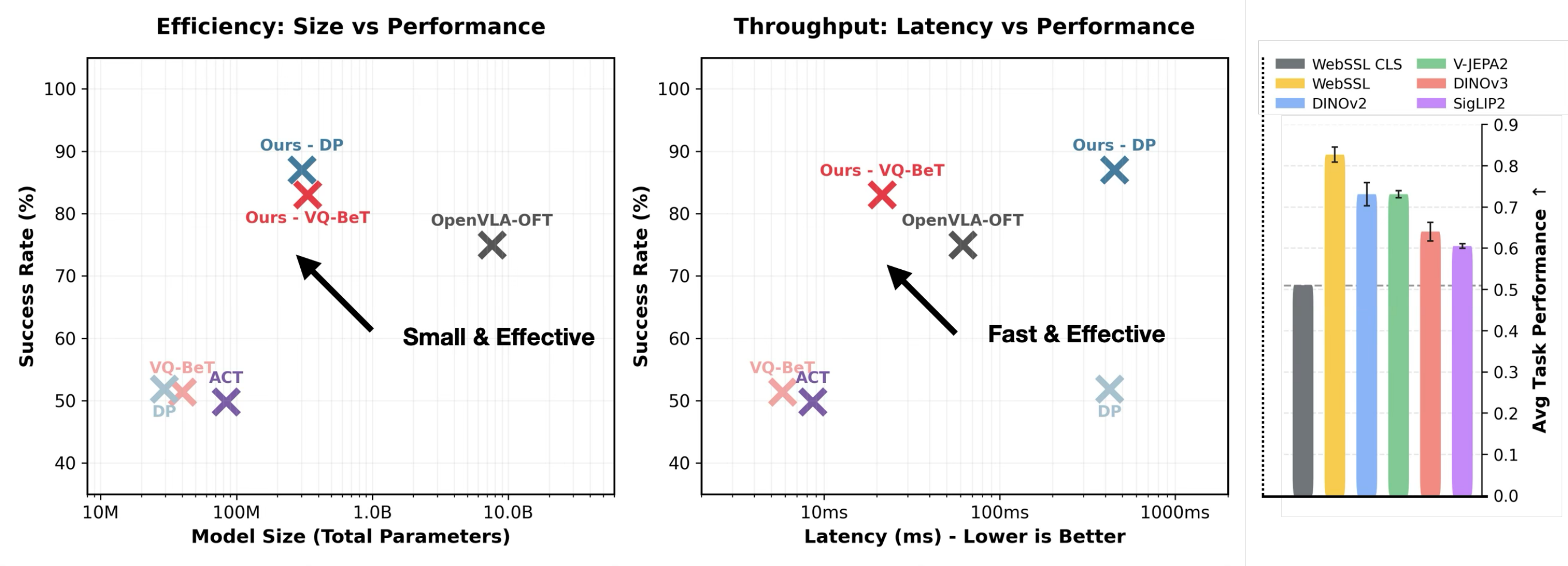}
\caption{We introduce \method{}, an efficient policy architecture that harnesses the power of pre-trained dense visual features. \method{} demonstrates superior performance while remaining computationally lean in both parameter count and inference latency. Through a rigorous analysis across five state-of-the-art visual representations, we show that our dense patch-based approach provides consistent performance gains over traditional global-feature policy baselines (WebSSL \texttt{CLS}). }
\label{fig:teaser}
\end{figure*}

Vision Transformers (ViTs) \cite{Dosovitskiy2020AnII} have become the de facto standard backbone for computer vision. By processing images as sequences of localized patches, ViTs extract rich, dense representations that preserve fine-grained spatial and semantic details.
This architectural shift, especially when combined with large-scale self-supervised and language-image pre-training \cite{He2021MaskedAA, Caron2021EmergingPI, Radford2021LearningTV, Chen2021AnES}, has driven state-of-the-art results across a vast array of tasks \cite{Radford2021LearningTV, Dosovitskiy2020AnII, dinov2, Liu2021SwinTH, liu2024grounding, siglip2, sam3, dinov3, webssl, vjepa, vjepa2}. Many of these visual capabilities, particularly fine-grained geometric understanding and robust feature localization, are directly useful for precise robotic manipulation.

Despite this clear advantage, the integration of dense ViT representations into robot learning has largely lagged behind, leaving the field bifurcated into two suboptimal paradigms. One approach, rooted in classical control and reinforcement learning, is to define the environment observation as a flat state vector. Visual observations are compressed into a global feature vector, derived from ResNet \cite{he2016deep} pooled features, or ViT \texttt{CLS} tokens. This approach creates a severe informational bottleneck: aggressive spatial compression destroys the fine-grained, local details necessary for precise manipulation. A newer approach is to fine-tune large vision-language models (VLMs) on action data. While these vision-language-action models (VLAs) have demonstrated strong performance across benchmarks, practitioners often have to fine-tune massive VLM backbones on their specific robot embodiments and downstream tasks for optimal performance, lugging around billions of parameters. Some optimization techniques improve the situation \cite{perez2018film, hu2022lora}, but training and inference remain quite expensive overall.

Can we inherit the representational gains of large-scale vision pretraining, without inheriting the cost of billion-parameter generative models? We argue for an efficient alternative: \textbf{visuomotor policies simply need dense features}. The dense visual understanding that makes VLAs effective is already present in Internet-scale pretrained ViTs, available off-the-shelf. By replacing global-pooled features with these patch features, \method{} captures the spatial detail relevant for precise manipulation at a fraction of the cost. We show that:
\begin{enumerate}[leftmargin=*,topsep=0pt,itemsep=-1ex,partopsep=1ex,parsep=1ex]
\item \textbf{Dense representations outperform global features for control}: on precise, multi-object/spatial tasks, spatially dense ViT patches substantially outperform global pooled features or \texttt{CLS} tokens while remaining competitive on the rest, regardless of the chosen policy architecture (\cref{tab:sim_results}, \cref{tab:real-results}, \cref{tab:egogym-results}).
\item \textbf{Pretrained ViT features transfer to control off-the-shelf}: frozen patch features from Internet-scale ViTs \cite{dasari2023datasets}, with no encoder fine-tuning, yield robust representations for control (\cref{tab:sim_results}, \cref{fig:vision}, \cref{tab:vision-vqbet}, \cref{tab:vision-dp}), and enable real-world precise manipulation (\cref{tab:real_robot_eval}, \cref{tab:real-results}).
\item \textbf{Spatial compression degrades control performance}: reducing spatial resolution, whether via pooling, or learned convolutional compression, degrades performance (\cref{tab:sim_results}, \cref{tab:compression}).
\item \textbf{\method{} is highly efficient}: \method{} matches or outperforms a large VLA fine-tuned on downstream tasks using 0.7\% of its parameter count, and runs at as low as $\sim11\text{ms}$ inference latency (\cref{tab:model-stats}, \cref{ssec:compute}).
\end{enumerate}

All code and data will be open-sourced. All hyperparameters and implementation details are in \cref{ssec:implementation} and \cref{sec:misc} for reproducibility.

\section{\method{}}

\begin{figure*}[t]
\center
\includegraphics[width=\textwidth]{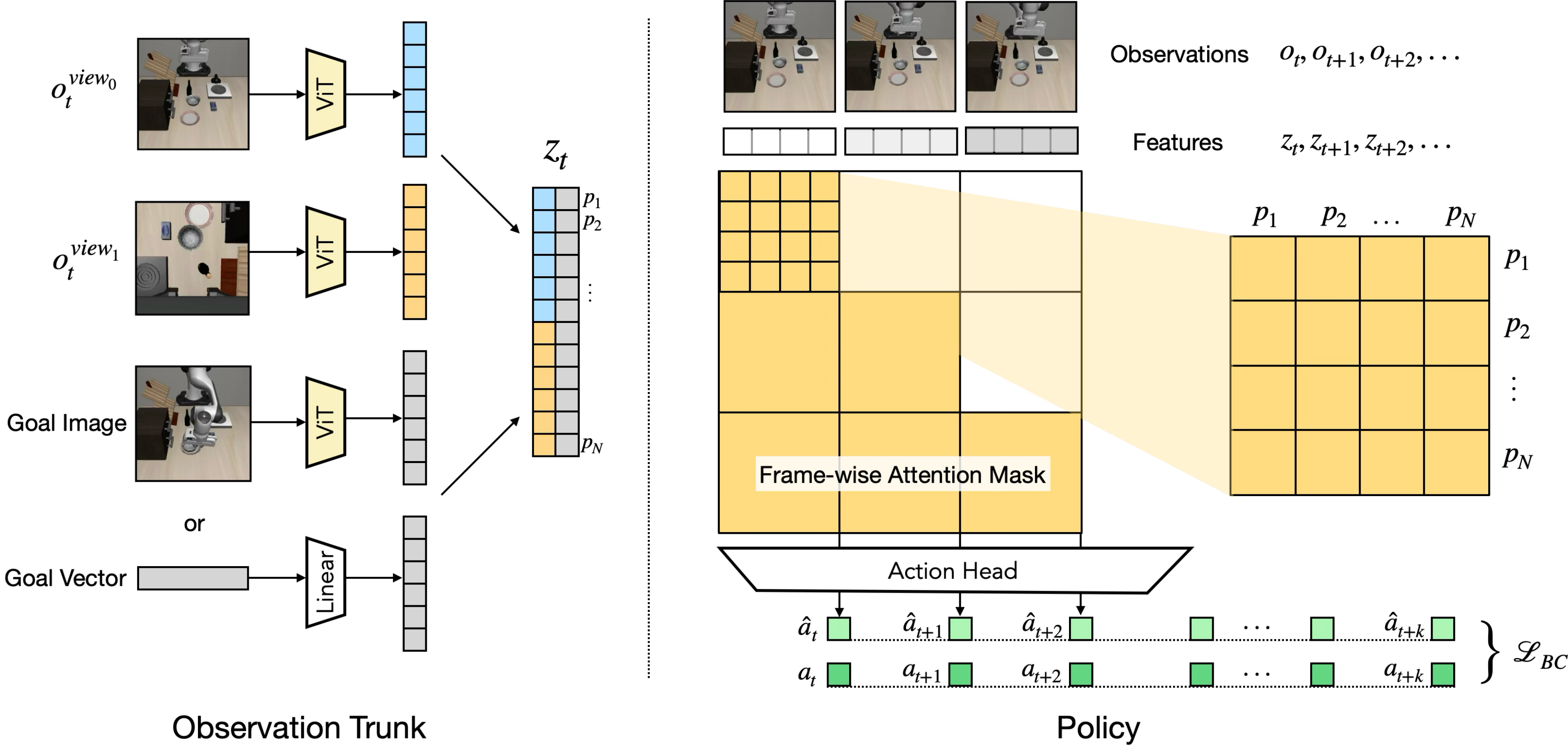}
\caption{\textbf{Architecture}. \method{} consists of an observation trunk (left) with a policy head (right). We encode multiview observations into patch features and optionally concatenate goal embeddings (images/states) into the current timestep sequence. \method{} is compatible with any transformer-based policy head. A block-wise causal mask is applied to enforce conditioning on past observations, allowing the policy to be optimized with any standard action head.}
\label{fig:method}
\end{figure*}

\method{} is a modular, Transformer-based~\citep{Vaswani2017AttentionIA} policy architecture designed to directly ingest uncompressed, dense patch features. It avoids the traditional global pooled feature bottleneck and preserves the fine-grained spatial information critical for precise manipulation. Our approach is compatible with a wide range of visual encoders and transformer-based policy architectures. The framework consists of an observation trunk (\Cref{sec:observation_trunk}), which encodes sensory inputs and goal specifications into a sequence of dense spatio-temporal features, and a policy head (\Cref{sec:policy}) that processes this sequence to emit actions.

\subsection{Observation Trunk}
\label{sec:observation_trunk}
Given an image observation $o_t \in \mathbb{R}^{C \times H \times W}$, a ViT encoder divides it into patches and extracts dense patch features of shape $P \times D$ (number of patches $\times$ patch embedding dimension). We simply take all patch features as the visual representation for downstream policy learning. For an observation context window of length $T$, the resulting feature is a sequence of patch features of shape $T \times P \times D$. This formulation is agnostic to the specific ViT architecture or pretraining objective, and is backwards-compatible with global pooled features or state-based environments by setting $P=1$.

For goal-conditioned behavioral cloning, we admit either a goal image or a goal vector input. For goals specified as an image, we encode it with the same encoder and concatenate it with the observation tensor to form a policy input tensor $z_t$ of shape $T \times P \times 2D$. For goals specified as a vector $g \in \mathbb{R}^{G}$, we concatenate it to every observation token to form a tensor of shape $T \times P \times (D+G)$. See \cref{fig:method} for a detailed illustration.

\subsection{Policy Learning}
\label{sec:policy}

We formulate policy learning as a sequence modeling problem over the extracted patch features. \method{} is compatible with any transformer-based policy architecture taking sequential inputs. To process the spatio-temporal patch feature tensor, we flatten the features into a sequence of length $T \times P$, add a learned 1D positional embedding indexed by the token's position in the flattened sequence, and apply a block-causal attention mask: patches maintain full bidirectional attention intra-frame but are causally masked inter-frame, allowing the model to integrate spatial information across each frame while preserving temporal causality, similar to~\cite{dinowm}. We ablate this choice in~\Cref{sec:attention_ablation}. Finally, the policy processes the masked sequence and emits a predicted action chunk at the last patch token for each frame with an action head. See \cref{fig:method} for details.

This formulation is agnostic to the action head architecture and training objective. In our experiments, we evaluate \method{} using two state-of-the-art architectures: Vector-Quantized Behavior Transformer (VQ-BeT) \cite{vqbet}, which uses a hybrid classification-regression loss, and Diffusion Policy (DP) \cite{diffusionpolicy}, which uses a denoising objective.

During training, we forward a sequence of patch tokens through the policy transformer trunk and action head, and compute a loss between the predicted and ground-truth actions for each frame. For inference, we extract the patch features from the current observation and append them to a rolling context window of length $T$. The policy predicts a chunk of actions from the observation context, and we execute the actions with receding horizon control.

\section{Experiments}

We evaluate \method{} across four simulated environments and three real robot manipulation environments, aiming to answer the following key research questions:

\begin{enumerate}[leftmargin=*,topsep=0pt,itemsep=-1ex,partopsep=1ex,parsep=1ex]
    \item How does \method{} compare to state-of-the-art policies using global and patch features?
    \item Does \method{} work for real-world precise manipulation?
    \item How do the latest pretrained encoders perform as representations for downstream policy learning?
    \item How does the spatial compression of visual features impact downstream task performance?
    \item How does \method{} compare to other methods in terms of efficiency? How do attention design and model size affect its performance? (Ablations \Cref{sec:ablations})
\end{enumerate}

\subsection{Environments}
\begin{figure}[t]
\center
\includegraphics[width=0.9\textwidth]{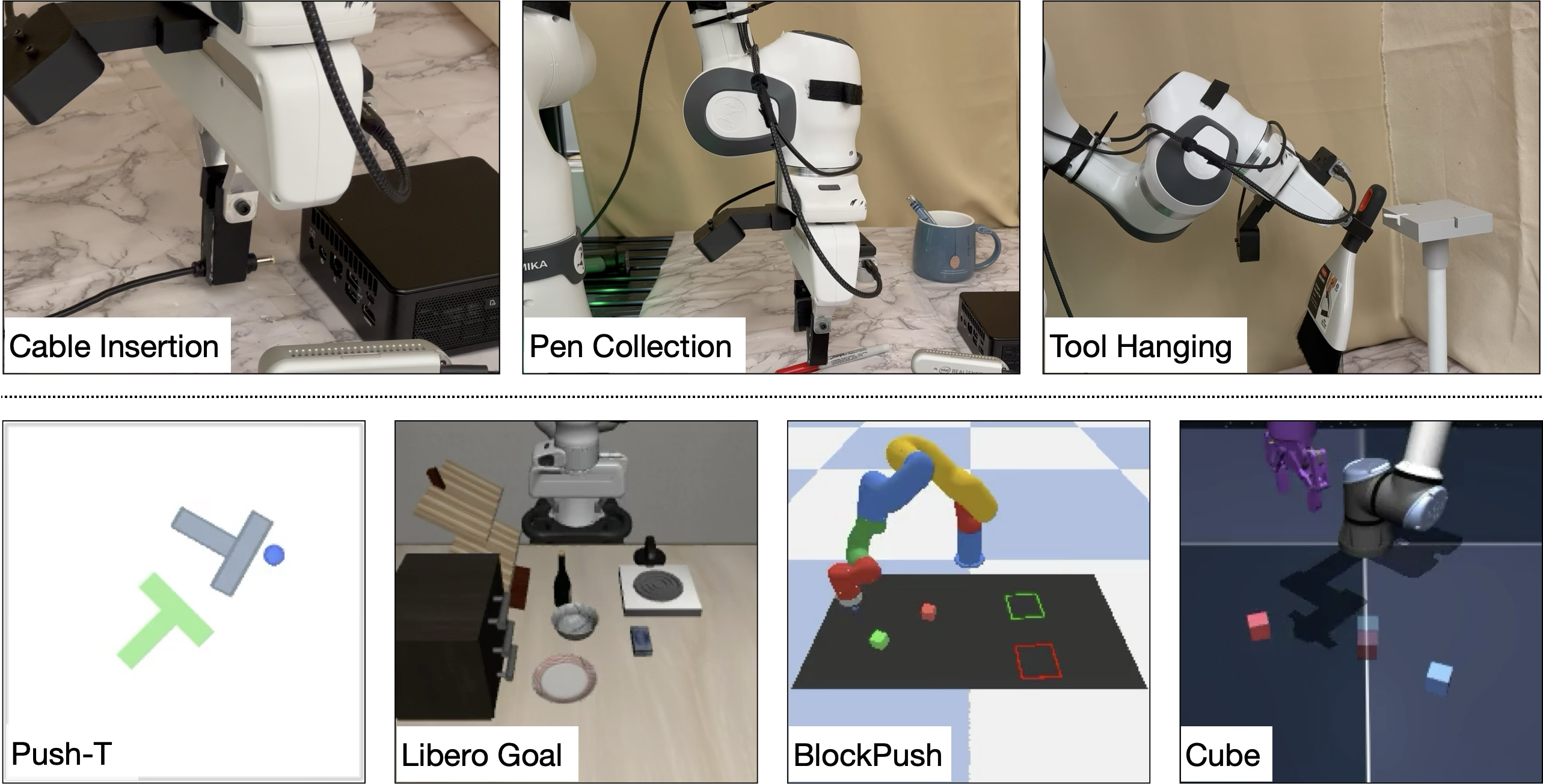}
\caption{We evaluate \method{} on four simulated and three real-world environments. }
\label{fig:envs}
\end{figure}

We evaluate \method{} across four simulated environments (Push-T, LIBERO Goal, BlockPush, Cube) with 2D-to-7D action spaces, and three real-world tasks using a 7-DoF Franka arm with a parallel-jaw gripper (inserting a power cable, hanging a tool, and collecting pens into a holder). Environments are visualized in \Cref{fig:envs} and detailed in App. \Cref{ssec:envs}.

\subsection{Baselines}
\label{sec:baselines}

We compare against two groups of baselines. To isolate the effect of patch features, we pair the same policy heads (VQ-BeT, Diffusion Policy) with standard global representations: DynaMo~\citep{dynamossl}, \texttt{CLS} tokens, and global average pooling. To position \method{} against the state of the art, we compare to ACT~\citep{Zhao2023LearningFB} and OpenVLA-OFT~\citep{openvla-oft}, which also consume dense features. Our baseline representations and policies are as follows:

\begin{enumerate}
    \item \textbf{Visual Representation Baselines:} To evaluate different feature extraction strategies, we compare against:
    \begin{itemize}
        \item \textbf{DynaMo~\cite{dynamossl}:} A global pooled representation learned via dynamics-based joint-embedding predictive architecture.
        \item \textbf{\texttt{CLS} Tokens:} The class token from the Vision Transformer, representing a compressed summary of the scene. 
        \item \textbf{Average Pooling:} A baseline that collapses the spatial feature map into a single vector via global average pooling. 
    \end{itemize}

    \item \textbf{ACT~\cite{Zhao2023LearningFB}: } A conditional VAE that predicts action chunks with temporal ensembling over overlapping chunks, reducing compounding error in fine manipulation. It uses patch features from a ResNet-18 vision encoder trained from scratch. 

     \item \textbf{OpenVLA-OFT~\cite{openvla-oft}:}  A Vision-Language-Action (VLA) model that finetunes OpenVLA with parallel action decoding and L1 action regression for faster training and inference. It builds on a Llama-2 7B LLM backbone and consumes channel-fused patch features from pretrained DINOv2~\cite{dinov2} and SigLIP~\cite{zhai2023sigmoid} vision encoders.
\end{enumerate}

\subsection{How well does \method{} work?}

\begin{table}[ht]
  \caption{\textbf{\method~on simulated environments.}
  \emph{Top}: standard policies that consume globally pooled visual features.
  \emph{Middle}: our pipeline with WebSSL patch features.
  \emph{Bottom}: other policies that consume patch tokens.
  \method{} substantially improves over global-feature counterparts of the same policies, while matching or exceeding patch-based baselines.}
  \label{tab:sim_results}
  \resizebox{\textwidth}{!}{%
  \begin{tabular}{llcccc}
  \toprule
  \textbf{Visual Representation} & \textbf{Policy} & \textbf{Push-T} & \textbf{LIBERO Goal} & \textbf{BlockPush} & \textbf{Cube} \\
  \midrule
  \multicolumn{6}{l}{\textit{Standard policies with globally pooled visual features}} \\
  \midrule
  DynaMo                   & VQ-BeT           & $0.66$                        & $0.93$                        & $0.65$                        & $0.28$                        \\
  WebSSL \texttt{Avg Pool} & VQ-BeT           & $0.54$\,\std{0.02}          & $0.97$\,\std{0.04}          & $0.84$\,\std{0.18}          & $0.25$\,\std{0.02}          \\
  WebSSL \texttt{CLS}      & VQ-BeT           & $0.59$\,\std{0.01}          & $0.95$\,\std{0.01}          & $0.77$\,\std{0.08}          & $0.23$\,\std{0.01}          \\
  DynaMo                   & Diffusion Policy & $0.73$                        & $0.68$                        & $1.06$\,\std{0.10}          & $0.27$                        \\
  WebSSL \texttt{Avg Pool} & Diffusion Policy & $0.79$\,\std{0.02}          & $0.98$\,\std{0.01}          & $1.34$\,\std{0.02}          & $0.21$\,\std{0.03}          \\
  WebSSL \texttt{CLS}      & Diffusion Policy & $0.68$\,\std{0.02}          & $\mathbf{0.99}$\,\std{0.01} & $0.99$\,\std{0.12}          & $0.21$\,\std{0.03}          \\
  \midrule
  \multicolumn{6}{l}{\textit{\textbf{\method{}}: patch features}} \\
  \midrule
  WebSSL Patch             & VQ-BeT \textbf{(Ours)}           & $0.68$\,\std{0.03}          & $0.94$\,\std{0.01}          & $\mathbf{1.68}$\,\std{0.15} & $1.68$\,\std{0.03}          \\
  WebSSL Patch             & Diffusion Policy \textbf{(Ours)} & $\mathbf{0.80}$\,\std{0.01} & $0.98$\,\std{0.00}          & $1.65$\,\std{0.08}          & $\mathbf{1.73}$\,\std{0.02} \\
  \midrule
  \multicolumn{6}{l}{\textit{Other patch-based policies (baselines)}} \\
  \midrule
  ResNet-18 Patch          & ACT          & $0.64$\,\std{0.03}          & $0.93$\,\std{0.02}          & $0.15$\,\std{0.01}          & $0.69$\,\std{0.11}          \\
  DINOv2 + SigLIP Patch    & OpenVLA-OFT  & $0.59$\,\std{0.02}          & $0.95$             & $1.43$\,\std{0.17}          & $1.50$\,\std{0.09} \\
  \bottomrule
  \end{tabular}}
  \end{table}

We evaluate \method{} on four simulated manipulation benchmarks (Push-T, LIBERO Goal, BlockPush, and Cube), spanning planar pushing, goal-conditioned multi-task manipulation, and multi-stage tasks. \Cref{tab:sim_results} organizes the comparison along the two axes that mirror our contribution: against the same policy classes consuming globally pooled visual features (top), and against other transformer-based policies that already consume patch tokens (bottom).

We report WebSSL~\citep{webssl} results in \cref{tab:sim_results} for its strong performance among the encoders we evaluate. A full comparison across encoders is provided in \cref{ssec:benchmark}. All policies receive only visual inputs. For the global representation baseline, we evaluate two variants commonly used in policy training: the \texttt{CLS} token and \texttt{Avg Pool} (global average pooling across all patch features). To maintain consistency across policies, we remove proprioceptive inputs from ACT's CVAE encoder, conditioning its latent variable $z$ on action chunk alone. We also add support for channel-stacking goal conditioning in ACT experiments, mirroring VQ-BeT and Diffusion Policy. We evaluate 100 trajectories per seed for each environment, reporting target coverage (Push-T), success rate (LIBERO Goal), or the average number of objects placed at the goal (BlockPush, Cube). We select the best-performing checkpoint for each run. For the OpenVLA-OFT baseline on LIBERO Goal, we report the results directly as provided in the original manuscript~\cite{openvla-oft}.

In \Cref{tab:sim_results}, \method{} using WebSSL patch features consistently outperforms global representations like DynaMo~\cite{dynamossl} on VQ-BeT and Diffusion Policy heads on precise, multi-object/spatial tasks (BlockPush, Cube), and competitive elsewhere. Replacing patches with a \texttt{CLS} token or average pooling (\texttt{Avg Pool}) severely degrades performance; the former favors high-level semantics over spatial details, while the latter lacks the resolution of uncompressed patches. Remarkably, \method{} outperforms the fine-tuned OpenVLA-OFT baseline which fuses both DINOv2 and SigLIP features on all four environments, confirming that dense spatial representations are vital for complex manipulation tasks.

\subsection{Real World Robotic Manipulation with \method{}}

We evaluate \method{} and baselines on three real-world manipulation tasks: Cable Insertion, Pen Collection, and Tool Hanging. These tasks are designed to test precise, long-horizon, and contact-rich manipulation, respectively. Detailed task descriptions can be found in \cref{ssec:envs}. We use DINOv2~\citep{dinov2} (ViT-S) patch features for all real-world experiments, a compact yet top-performing backbone in our representation study (\cref{ssec:benchmark}) that is well established for strong visual understanding~\cite{dinov2, dinowm}. Additionally, in \cref{ssec:cap-results} and \cref{ssec:egogym}, we evaluate a zero-shot general object pickup policy trained with our method, and compare with the original global-pooled policy following the setup of Contact-Anchored Policies \cite{cui2026contact}.

\begin{figure*}[ht]
    \centering
    \includegraphics[width=\textwidth]{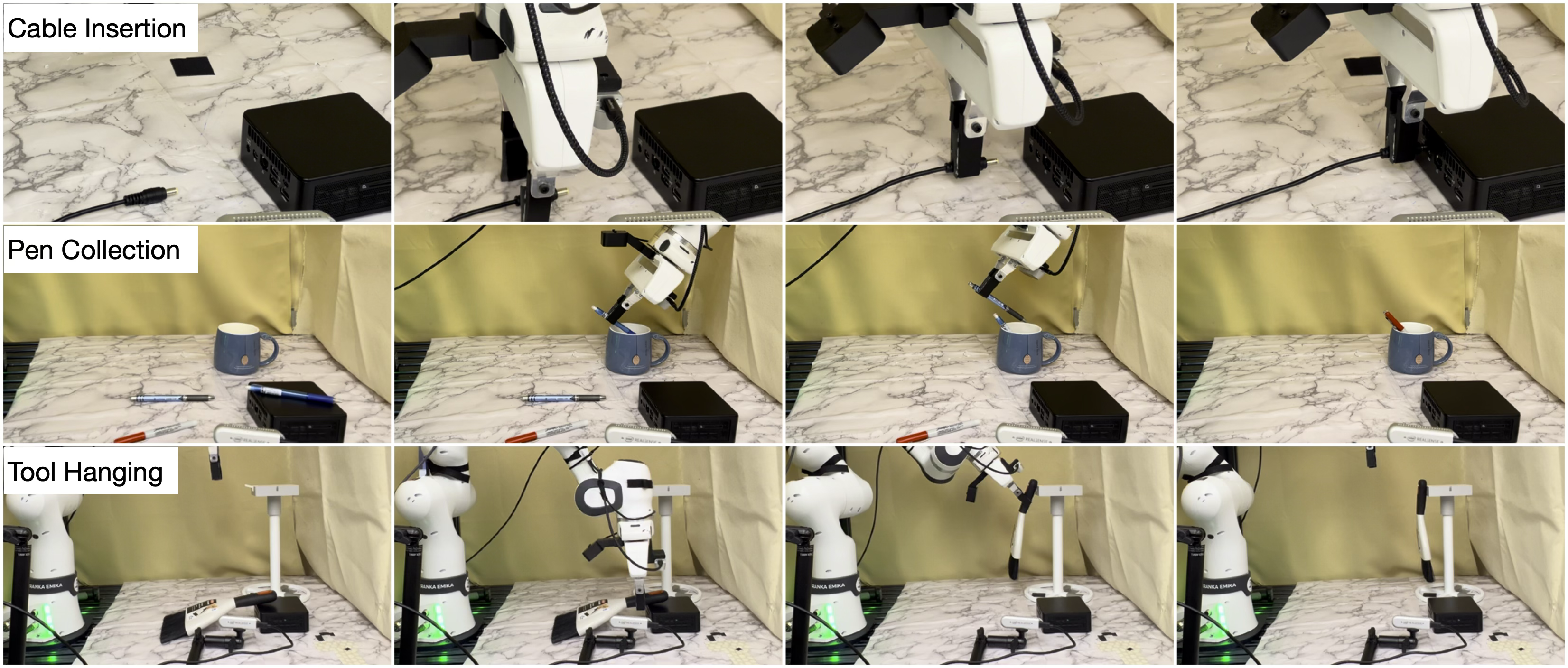}
    \caption{
    Real-robot rollout examples for the three evaluation tasks: cable insertion, pen collection, and tool hanging.
    Each row shows one task rollout and visualizes key stages of the policy execution.}
    \label{fig:rollouts}
\end{figure*}

Table~\ref{tab:real_robot_eval} reports cumulative success rates through each task stage. Nearly every method completes the initial grasp; as we progress into the task, however, we find that \method{} outperforms baselines on every task (visualizations in \cref{fig:rollouts}). The improvement is most significant on the lowest-tolerance task of Cable Insertion, where the barrel jack must be inserted with a $\sim$2\,mm tolerance. The improvements over global pooled features mirror our simulation findings. Notably, \method{} outperforms a fine-tuned OpenVLA-OFT on all three tasks, showing that for in-domain task learning, a lightweight policy trained from scratch over frozen dense features can be competitive with a heavy pretrained VLA fine-tuned on the same data. In Cable Insertion, OpenVLA-OFT often gets stuck near the socket, suggesting that these failures are dominated by fine-grained closed-loop control after contact, rather than task-level semantic understanding.

\begin{table}[h]
\centering
\small
\caption{Real-robot success rates through task stages, 20 trials.}
\label{tab:real_robot_eval}
\resizebox{\textwidth}{!}{%
\begin{tabular}{llccc}
\toprule
\textbf{Visual Representation} & \textbf{Policy} & \multicolumn{3}{c}{\textbf{Task Stage}} \\
\midrule
\multicolumn{2}{l}{\textit{\textbf{Cable Insertion}}} & Picked up cable & Plugged into port & Fully inserted \\
\cmidrule(lr){3-5}
DINOv2 Patch          & VQ-BeT \textbf{(Ours)} & \textbf{1.00} & \textbf{0.85} & \textbf{0.70} \\
DINOv2 \texttt{CLS}   & VQ-BeT                 & \textbf{1.00} & 0.70          & 0.60          \\
ResNet-18 Patch       & ACT                    & \textbf{1.00} & 0.40          & 0.35          \\
DINOv2 + SigLIP Patch & OpenVLA-OFT            & \textbf{1.00} & 0.55          & 0.30          \\
\midrule
\multicolumn{2}{l}{\textit{\textbf{Pen Collection}}} & First pen placed & Second pen placed & Third pen placed \\
\cmidrule(lr){3-5}
DINOv2 Patch          & VQ-BeT \textbf{(Ours)} & \textbf{1.00} & \textbf{1.00} & \textbf{0.85} \\
DINOv2 \texttt{CLS}   & VQ-BeT                 & \textbf{1.00} & 0.95          & 0.65          \\
ResNet-18 Patch       & ACT                    & \textbf{1.00} & 0.85          & 0.65          \\
DINOv2 + SigLIP Patch & OpenVLA-OFT            & \textbf{1.00} & 0.85          & 0.60          \\
\midrule
\multicolumn{2}{l}{\textit{\textbf{Tool Hanging}}} & Picked up & Reached hook & Tool placed \\
\cmidrule(lr){3-5}
DINOv2 Patch          & VQ-BeT \textbf{(Ours)} & \textbf{1.00} & \textbf{0.90} & \textbf{0.90} \\
DINOv2 \texttt{CLS}   & VQ-BeT                 & \textbf{1.00} & 0.75          & 0.70          \\
ResNet-18 Patch       & ACT                    & \textbf{1.00} & 0.85          & 0.85          \\
DINOv2 + SigLIP Patch & OpenVLA-OFT            & 0.95          & \textbf{0.90} & 0.65          \\
\bottomrule
\end{tabular}}
\end{table}

\subsection{Benchmarking Pre-trained Visual Representations}
\label{ssec:benchmark}

While \method{} achieves superior results using WebSSL patch features, it remains to be seen how alternative pre-trained representations impact downstream policy learning. To investigate the extent to which the choice of vision backbone influences performance, we evaluate five state-of-the-art visual representations: DINOv2~\cite{dinov2}, DINOv3~\cite{dinov3}, WebSSL~\cite{webssl}, V-JEPA~2~\cite{vjepa2}, and SigLIP~2~\cite{siglip2}. Detailed descriptions of each encoder are provided in~\Cref{sec:visual_rep_intro}.

\begin{figure*}[h]
\center
\includegraphics[width=\textwidth]{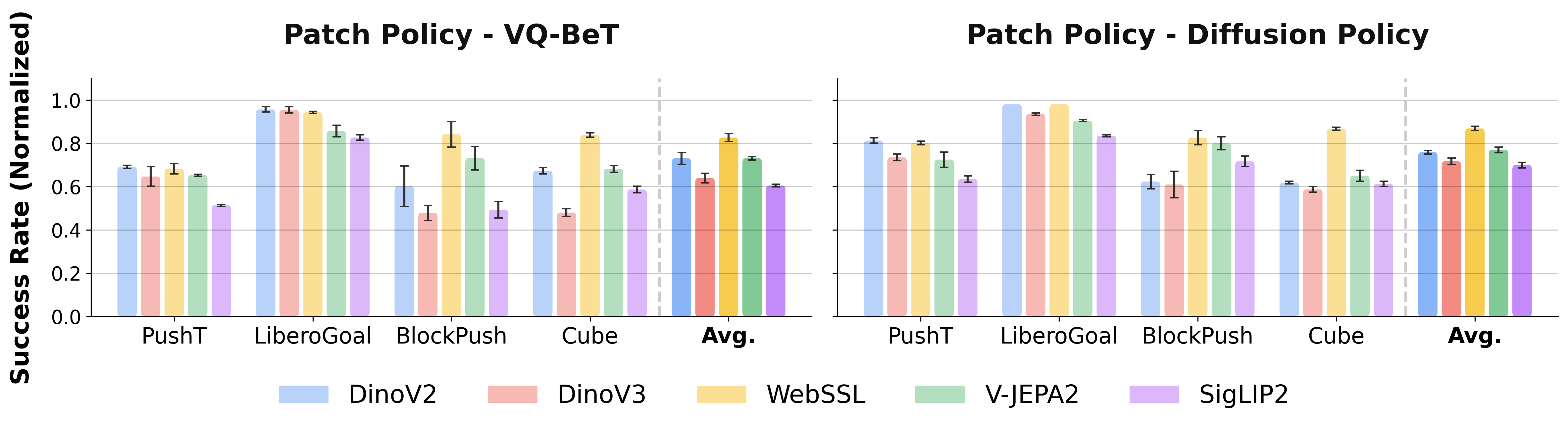}
\caption{\textbf{Comparison of \method{} across various pretrained visual representations}. We report the mean and standard deviation across runs with three random seeds. Our results suggest that DINOv2 and WebSSL are the most effective vision backbones for robot learning tasks. We normalize BlockPush and Cube results to $[0, 1]$ to facilitate comparison across the entire task suite.}
\label{fig:vision}
\end{figure*}

For all experiments, we freeze the image encoder and exclusively train the policy. This approach isolates and evaluates the out-of-the-box representation capabilities of each backbone. This frozen setup also allows us to precompute the visual embeddings, significantly accelerating training.

We report the performance on the simulated environments in~\Cref{fig:vision} and full results in~\Cref{tab:vision-vqbet} and \Cref{tab:vision-dp}, as well as the average performance across all tasks per policy backbone. As illustrated in \Cref{fig:vision}, WebSSL and DINOv2 achieve the highest performance across the majority of tasks. This establishes them as robust visual backbones for robotic control and aligns with previous findings in world modeling~\cite{Goswami2025WorldMC}. SigLIP~2 falls short across the environments. We hypothesize that SigLIP~2's emphasis on semantic language-image alignment sacrifices the dense geometric features necessary for manipulation. Since our tasks prioritize spatial reasoning over linguistic understanding, this vision-language grounding provides a less effective signal for policy learning. 

Notably, the relative ranking of visual representations remains remarkably consistent across different policy architectures for any given task. This consistency suggests that the quality of the visual representation is still a primary bottleneck for policy learning, independent of the downstream action head. Furthermore, the average ranking of these representations remains stable across the entire task suite, underscoring the generalizability of certain pre-training objectives. Based on these findings, we recommend the use of WebSSL or DINOv2 as the vision backbones for robot learning tasks.

\subsection{Is \method{} computationally efficient?}
\label{ssec:compute}
Beyond task performance, we evaluate the computational feasibility of \method{} for real-time robotic applications. In \Cref{tab:model-stats}, we compare the parameter counts and inference latency (ms) of \method{} variants against baseline policies on an NVIDIA H200 GPU. 

\begin{table}[ht]
\centering
\caption{Computational Resources and Inference Speed of \method{}.}
\label{tab:model-stats}

\setlength{\tabcolsep}{4pt} 
\renewcommand{\arraystretch}{1}

  \begin{tabular}{lccc}
  \toprule
  \textbf{Method} & \textbf{Total Params} & \textbf{Trainable Params} & \textbf{Inference Latency (ms)} \\
  \midrule
  VQ-BeT (ResNet-18) & 39.95M & 28.77M & 5.79   \\
  Ours - VQ-BeT (DINOv2) & 51.55M & 29.49M & 10.99   \\
  Ours - VQ-BeT (WebSSL) & 334.00M & 30.34M & 21.43   \\
  DP (ResNet-18)     & 29.35M & 9.09M  & 421.89 \\
  Ours - DP (DINOv2)     & 40.43M & 9.19M  & 445.85 \\
  Ours - DP (WebSSL)     & 303.66M & 9.35M  & 451.68 \\
  \midrule
  OpenVLA-OFT             & 7.61B  & 177.90M & 61.71  \\
  ACT             & 83.85M & 83.85M & 8.63   \\
  \bottomrule
  \end{tabular}
  \end{table}

\textbf{Parameter Efficiency.}  \method{} beats OpenVLA-OFT with under 5\% of its parameters with ViT-L (as little as $\sim$0.7\% with ViT-S), showing that what matters is not scale but direct access to dense patch features from a frozen, Internet-pretrained backbone. \Cref{sec:misc} has training details and model size ablations.

\textbf{Inference Latency.} Real-time control typically requires high-frequency feedback loops. Here we report the raw model inference latency, defined as the time required for a single forward pass which does not include temporal speedups from action chunking, as such techniques can be applied orthogonally to all evaluated methods. All evaluations are on a single H200 GPU. Our VQ-BeT variants demonstrate exceptional speed even when processing dense DINOv2 patch features (10.99~ms), comparable to ACT with ResNet-18 at 8.63~ms. In contrast, while OpenVLA-OFT is optimized for efficiency, it still requires 61.71~ms per forward pass. We observe that switching from global representations (ResNet) to patch features (DINOv2) in \method{} results in a predictable increase in latency due to the longer token sequence (5.79~ms $\rightarrow$ 10.99~ms for VQ-BeT); however, the resulting latency remains well within the requirements for high-speed manipulation. Notably, the bottleneck for Diffusion Policy variants remains the iterative denoising process rather than the visual representation, as evidenced by the negligible difference in inference latency between the global (421.89~ms) and patch-based (445.85~ms) DP variants.

\textbf{Training Cost.} \method{} with DINOv2 patch features converges in 6.5 hours on 1xL40S (6.5 GPU-hours); OpenVLA-OFT converges in 4 hours on 4x L40S (16 GPU-hours); ACT converges in 12 hours on 2xL40S (24 GPU-hours). This efficiency reflects both the small trainable parameter count and the frozen pretrained visual backbone.

\subsection{Should we compress patch features?}

\begin{wraptable}{r}{0.32\textwidth} 
\centering
\vspace{-15pt} 
\caption{Patch Compression}
\label{tab:compression}
\small 
\begin{tabular}{lc}
\toprule
\textbf{Resolution} & \textbf{Push-T} \\
\midrule
256 patches & \textbf{0.69} \\
64 patches  & 0.52 \\
16 patches  & 0.53 \\
4 patches   & 0.51 \\
1 patch     & 0.48 \\
\bottomrule
\end{tabular}
\end{wraptable}

\method{} achieves strong performance using raw patch features. A natural question is whether patch tokens for each frame can be compressed to improve efficiency without sacrificing policy performance. We introduce a lightweight convolutional encoder, trained end-to-end with the policy, to compress the frozen DINOv2 patch features spatially, reducing the number of patch tokens per frame. On Push-T (\Cref{tab:compression}), spatially downsampling the features results in a significant decrease in task success. This trend confirms that the fine-grained spatial density of the original tokens is crucial for precise control. While compression remains a viable trade-off for strictly hardware-constrained deployments, we recommend preserving the uncompressed patch resolution whenever compute permits.

\subsection{Additional Ablations}
\Cref{sec:ablations} contains more experiments and details on attention masks, model size, and visualizations.

\section{Related Work}

\subsection{Imitation Learning}

Imitation Learning (IL) enables agents to learn skills from expert demonstrations without explicit reward engineering \cite{pomerleau1988alvinn}. While early IL focused on low-dimensional states, recent paradigms shift toward vision-based policies operating on high-dimensional inputs. Despite advances in expressive architectures and generative objectives that model multimodal actions from large datasets~\cite{diffusionpolicy, Zhao2023LearningFB, vqbet, Shafiullah2022BehaviorTC, Cui2022FromPT, florence2022implicit, rana2025imle, OpenXR, khazatsky2024droid}, a performance gap persists between state-based and vision-based agents. \method{} addresses this within vision-based behavioral cloning. Rather than proposing a novel training objective, we close this gap by equipping standard architectures with the dense visual representations well-established in computer vision.

\subsection{Visual Representation for Embodied Learning}

Visual representation for control has rapidly evolved from in-domain self-supervised methods~\cite{moco, Grill2020BootstrapYO, dynamossl} to Vision Transformers (ViTs)~\cite{Dosovitskiy2020AnII}—such as DINO~\cite{dino, dinov2, dinov3}, V-JEPA~\cite{vjepa, vjepa2}, and SigLIP~\cite{zhai2023sigmoid, siglip2}—that extract dense patch features instead of compressed global vectors~\cite{r3m}. Despite their success in world modeling~\cite{dinowm, Goswami2025WorldMC, vjepa2}, their adoption in robotic learning remains limited. Because standard continuous control backbones assume compressed global states~\cite{florence2022implicit, Shafiullah2022BehaviorTC, diffusionpolicy, vqbet}, most architectures still rely on global pooling. Aside from computationally heavy VLA models~\cite{Brohan2023RT2VM}, efficient robotic agents rarely process uncompressed spatial tokens. \method{} bridges this gap by directly integrating foundational patch features into lightweight policies.

\subsection{Transformer-based Policies and VLAs}
Transformers~\cite{Vaswani2017AttentionIA} have achieved remarkable success across NLP~\cite{Devlin2019BERTPO, brown2020language} and vision~\cite{Dosovitskiy2020AnII, Liu2021SwinTH, vjepa2}. Its capacity for long-range dependency modeling makes it natively suitable for policy learning~\cite{Jaegle2021PerceiverGP}. Transformer-based policies have been widely adopted in both RL~\cite{Chen2021DecisionTR, Janner2021OfflineRL} and IL~\cite{Shafiullah2022BehaviorTC, Cui2022FromPT, Zhao2023LearningFB, su2025dense, baku, vqbet}. These models typically assume a compressed global feature vector for vision, discarding spatial granularity. While some agents do utilize multiple spatial tokens~\cite{Zhao2023LearningFB, rt1, Reed2022AGA, hou2024diffusion, Team2024OctoAO, goyal2023rvt, shridhar2023perceiver}, they often rely on end-to-end encoder training or finetuning in-domain instead of using pretrained vision backbones. We include ACT~\cite{Zhao2023LearningFB} as a representative baseline from this category, which operates on patch tokens from a ResNet-18 encoder trained from scratch. One prominent class of methods that utilizes dense features is the Vision-Language-Action (VLA) model, which jointly learns language understanding, control, and perception~\cite{Brohan2023RT2VM, OpenXR, Driess2023PaLMEAE, pi0, pi05, pi06} by fine-tuning large Vision-Language Models (VLM) on large-scale embodied data. The large backbone introduces latency for real-time control, which recent optimizations such as OpenVLA-OFT~\cite{openvla-oft} begin to address. We adopt OpenVLA-OFT as our primary VLA baseline. \method{} disentangles the utilization of dense visual representations from general language grounding and massive model scale, and provides a lightweight pipeline that directly ingests these dense features for imitation learning.

\section{Limitations}

While \method{} effectively leverages dense spatial features for control, several directions remain for future work. First, we focused exclusively on frozen vision backbones, and future work could explore end-to-end fine-tuning to adapt these representations for specialized visual domains. Second, dense tokens increase sequence length and training time. Optimizations like FlashAttention \cite{dao2022flashattention} could accelerate both training and inference. Finally, \method{} is currently evaluated purely as a behavior cloning policy. Extending this patch-based architecture to reinforcement learning could be a promising direction to surpass the performance ceiling of static expert demonstrations.

\section{Conclusion}

In this work, we investigate the efficacy of patch-level features for robot learning and introduce \method{}, an efficient policy class that harnesses pre-trained dense visual representations. It achieves superior performance while remaining computationally lean across parameter count, training compute, and inference latency, outperforming both global-feature policies and heavyweight VLAs. By keeping the visual backbone frozen and ingesting its patch tokens directly through a block-causal attention mask, \method{} provides the robotics community with a lightweight, drop-in pipeline to readily absorb continuing progress in visual representation learning, without sacrificing the fast training and high-frequency control that real-world manipulation demands.


\acknowledgments{We would like to thank Mahi Shafiullah, Irmak Guzey, Kevin Wu, Hengkai Pan, Zifan Zhao, Alex Xiaole Jiang, Andre Wang, Zicheng Teng, Kanad Patel, Yvonne Wu, Zavier Andrianarivo for their valuable discussion and feedback. This work was supported by grants from LG, Qualcomm, Honda, Microsoft, Hyundai, NSF award 2339096 and ONR awards N00014-21-1-2758 and N00014-22-1-2773, and AFOSR under grant FA95502310139. Lerrel Pinto is supported by the Sloan, Packard, and CIFAR Fellowships.}


\bibliography{main}

\newpage
\appendix
\section{Appendix}
\subsection{Implementation and Baselines}
\label{ssec:implementation}

Our implementation and baselines are built upon the following open-source repositories:

\begin{enumerate}
  \item DynaMo:     \url{https://github.com/jeffacce/dynamo_ssl}
  \item VQ-BeT:   \url{https://github.com/jayLEE0301/vq_bet_official}
  \item Diffusion Policy: \url{https://github.com/real-stanford/diffusion_policy}
  \item OpenVLA-OFT:      \url{https://github.com/moojink/openvla-oft}
  \item ACT: \url{https://github.com/tonyzhaozh/act}
\end{enumerate}

\subsection{Environments and Tasks}
\label{ssec:envs}
We evaluate \method{} across four simulated environments (Push-T, LIBERO Goal, BlockPush, Cube) with 2D-to-7D action spaces, and three real-world tasks using a 7-DoF Franka arm with a parallel-jaw gripper. The real-world tasks comprise: (1) \textit{Tool Hanging}, (2) \textit{Pen Collection} (long-horizon, small objects), (3) \textit{Cable Insertion} (low-tolerance insertion). We additionally test the generalization capabilities of \method{} in another real robot task: \textit{Franka Pickup} (unseen object generalization). For each environment, we detail the experimental setup, dataset size and collection process, and the metrics for evaluation. Refer to \Cref{fig:envs} for environment visualizations.

\begin{enumerate}
    \item \textbf{Push-T~\cite{diffusionpolicy}: } This environment requires a pusher agent with 2-dimensional action space to maneuver a green T-shaped block into a fixed, gray T-shaped target zone. Observations consist of a shape $224\times224$ top-down image of the scene. The dataset has 206 expert trajectories from Diffusion Policy~\cite{diffusionpolicy}. We evaluate over 100 rollouts and report the final spatial coverage proportion of the green block over the target (maximum 1.0). See \Cref{fig:sim-evals} for a sample rollout of \method{} on Push-T. 
    \item \textbf{LIBERO Goal~\cite{liu2023libero}: } This environment suite from the LIBERO benchmark consists of 10 manipulation tasks in one scene with a 7-DoF Franka arm. 

    Observations are multi-view, providing shape $224\times224$ images from both a third-person and a wrist-mounted camera. The dataset contains 50 expert trajectories per task, yielding 500 total demonstrations. We evaluate the suite across 100 trials (10 per goal) and report the overall success rate (maximum 1.0). See \Cref{fig:sim-evals} for a sample rollout of \method{} for each task.

    \item \textbf{BlockPush~\cite{florence2022implicit}: } This environment features two blocks and two square target zones. A robotic pusher, controlled via 2D end-effector translation, must push both blocks into designated targets (either same- or opposite-colored). The state is captured via two $224\times224$ third-person camera views. We train on 1000 expert demonstrations and evaluate across 100 rollouts, reporting the mean number of correctly placed blocks (maximum 2.0).
    
    \item \textbf{Cube~\cite{ogbench}:} This environment from OGBench~\cite{ogbench} features a 6-DoF UR5e robot arm interacting with two cube blocks in a pick-and-place task. The goal is for the arm to stack cubes into a fixed designated configuration at the center of the plane. The observation is a $224\times224$ image of a third-person view of the scene. The initial positions of the cube blocks are randomized. For training, we generate a dataset of 550 successful trajectories, where success requires correct stacking order and placement within 200 steps. The dataset was collected by an expert policy with temporally correlated noise following OGBench~\cite{ogbench}. We evaluate 100 rollouts and report the mean number of blocks reaching target locations (maximum 2.0). See \Cref{fig:sim-evals} for a sample rollout of \method{} for each task.

    \item \textbf{Tool Hanging:} In the tool hanging task, a brush is initially placed on the table. The robot must pick up the brush and hang it on a small hook mounted on a stand. A trial is considered successful if the robot hangs the brush stably on the hook. This task requires grasping the tool from the table, accurately positioning it relative to a small hook, and achieving a stable final configuration upon release. We use two wrist-mounted cameras, one facing forward and one facing downward. We collect 50 demonstration episodes, corresponding to approximately 18 minutes of real-robot data.

    \item \textbf{Pen Collection:} In the pen collection task, the robot must pick up three pens and place all of them into a pen holder. Each trial contains three different pens and three possible initial positions. The pens are not assigned to fixed positions and can be permuted across trials. The initial pose of each pen is randomized within approximately $\pm 1$ cm in position and $\pm 15^\circ$ in orientation. This task evaluates multi-object manipulation, robustness to object-pose variation, and long-horizon execution over repeated pick-and-place stages. We use two wrist-mounted cameras, one facing forward and one facing downward, to provide complementary visual observations during grasping and placement. We collect 52 demonstration episodes, corresponding to approximately 45 minutes of real-robot data.

    \item \textbf{Cable Insertion:} In the cable insertion task, the robot is required to pick up a power cable and insert it into a NUC. The NUC position is fixed across all demonstrations and evaluation trials, while the initial pose of the cable is randomized within a $2$cm $\times$ $2$cm square. The final insertion tolerance is around $2$mm. This task requires the robot to first grasp a barrel jack, align the connector with the port, and complete a precise insertion. To ensure that the insertion process is fully observable, we use two camera views: a wrist-mounted camera and a side-view camera. We collect 101 demonstration episodes, corresponding to approximately 37 minutes of real-robot data.
    
    \item \textbf{CAP - EgoGym~\cite{cui2026contact}:} Introduced in Contact-Anchored Policies (CAP) \cite{cui2026contact}, this simulated benchmark suite consists of three zero-shot generalization environments: general object pickup, door/drawer opening, and door/drawer closing. The dataset contains 14,606 real-world demonstrations for pickup, 3,690 for opening, and 2,069 for closing, collected via a handheld tool. \Cref{fig:cap-dataset-samples} shows sample trajectories from the dataset for each of the tasks. The simulation procedurally generates scenes with diverse textures to test zero-shot generalization. \Cref{fig:egogym-evals} shows \method{} rollouts in EgoGym. 

    \item \textbf{CAP - Franka Pickup:} For real world evaluations, we set up a physical Franka FR3 robot and ten evaluation objects exactly following the hardware and object configuration established in CAP~\cite{cui2026contact}. The observation is a $224\times224$ image from a wrist-mounted camera, and a 3D point specifying the expected physical contact location. For each evaluation, we evaluate with five objects in scene, and condition the policy to pick up each of the five objects for ten trials, for the two sets of five objects, for a total of 100 evaluations. See \Cref{fig:cap-success} and \Cref{fig:cap-failure} for rollout trajectories of \method{}.

\end{enumerate}

\subsection{Zero-shot Manipulation in the Real World}
\label{ssec:cap-results}
\begin{figure}[ht]
    \centering
    \includegraphics[width=\textwidth]{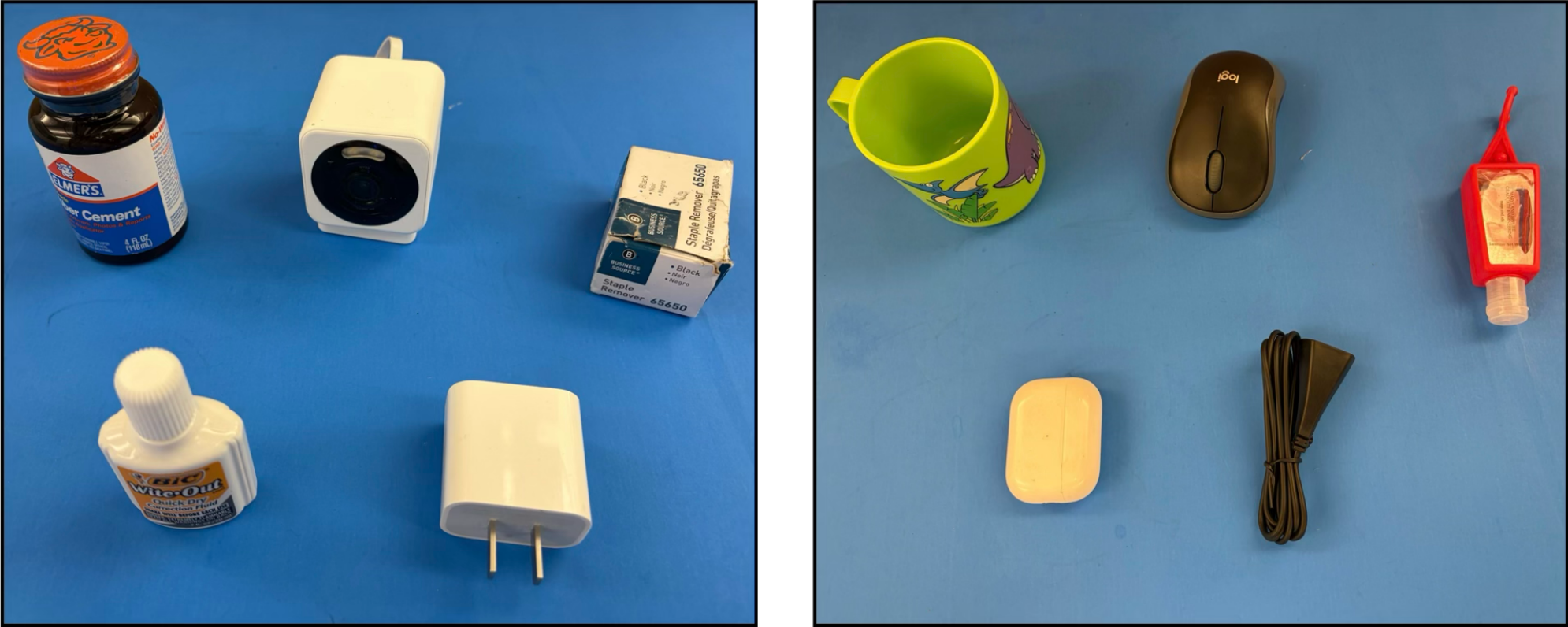} 
    \caption{We evaluate on 10 unseen objects for Real Franka Pickup, following the evaluation criteria in CAP \cite{cui2026contact}.}
    \label{fig:real-pickup-viz}
\end{figure}

To evaluate whether gains from \method{} apply in the zero-shot generalization domain, where the policy is trained on a large-scale real-world dataset, and evaluated in out-of-distribution environments and on novel objects (\Cref{fig:real-pickup-viz}), we train a VQ-BeT policy on the CAP object pickup dataset with DINOv2 patch features. We evaluate it on the real Franka robot environment, where the robot attempts to pick up 10 objects unseen during training for a total of 100 trials (\Cref{ssec:envs}). We compare it with the original CAP checkpoint with global-pooled features on the CAP gripper embodiment, following the CAP evaluation criteria (\Cref{sec:baselines}). We see that \method{} improves over the vanilla CAP checkpoint in zero-shot object pickup.

\begin{table}[ht]
    \caption{Real-world zero-shot object pickup results.}
    \centering
    \label{tab:real-results}

    \setlength{\tabcolsep}{5pt} 
    \renewcommand{\arraystretch}{1}
    \begin{tabular}{lc}
    \toprule
    \textbf{Method} & \textbf{Real Franka Pickup} \\
    \midrule
    CAP             & 79\%                    \\
    Ours            & \textbf{87\%}          \\
    \bottomrule
    \end{tabular}
\end{table}

\Cref{fig:cap-success} shows rollout success trajectories for each object, and \Cref{fig:cap-failure} shows rollout failure modes (early grasping, and overshooting).

\subsubsection{Zero-shot real-to-sim evaluations.}
\label{ssec:egogym}
Additionally, CAP \cite{cui2026contact} and SIMPLER \cite{li24simpler} have shown that for policies trained on real-world datasets, simulation evaluations can be highly predictive of real-world performance. To this end, we train patch-based policies on object pickup, door/drawer opening, and closing data, and compare them with the released global feature CAP checkpoints, evaluating both on diverse background textures and objects for 5000 episodes in EgoGym. We see in \Cref{tab:egogym-results} that patch policies outperform global feature policies across all three tasks in the EgoGym benchmark in diverse evaluations. Rollout trajectory samples are shown in \Cref{fig:egogym-evals}.

\begin{table}[ht]
\centering
\caption{Real-to-sim evaluations on EgoGym.}
\label{tab:egogym-results}

\setlength{\tabcolsep}{5pt} 
\renewcommand{\arraystretch}{1}

\begin{tabular}{lccc}
\toprule
\textbf{Method} & \textbf{EgoGym Pickup} & \textbf{EgoGym Open} & \textbf{EgoGym Close} \\
\midrule
CAP             & 75.78\%                   & 67.88\%                 & 86.50\%  \\
Ours            & \textbf{79.50\%}          & \textbf{71.40\%}        & \textbf{92.44\%} \\
\bottomrule
\end{tabular}
\end{table}

\subsection{Benchmarking Pretrained Visual Representations for Policy Learning}
\label{sec:visual_rep_intro}
While \method{} achieves superior results using DINOv2 patch features, it remains to be seen how alternative pre-trained representations impact downstream policy learning. To investigate the extent to which the choice of vision backbone influences performance, we evaluate \method{} across the following state-of-the-art visual representations:

\begin{enumerate}
    \item \textbf{DINOv2~\cite{dinov2}:} A Vision Transformer pre-trained on a curated dataset of 142M images (\texttt{LVD-142M}) using discriminative and reconstructive self-supervised losses; for a $224 \times 224$ input, we use the ViT-S/14 backbone yielding $16 \times 16 \times 384$ patch features.
    
    \item \textbf{DINOv3~\cite{dinov3}:} An evolution of the DINO framework scaling to larger, internet-scale datasets with improved stability; for a $224 \times 224$ input, we employ the ViT-S/16+ variant providing $14 \times 14 \times 384$ patch features.
    
    \item \textbf{WebSSL~\cite{webssl}:} A model that applies the DINO self-supervised loss to massive, uncurated web-crawled datasets; for a $224 \times 224$ input, the encoder outputs $16 \times 16 \times 1024$ patch features.
    
    \item \textbf{V-JEPA~2~\cite{vjepa2}:} A Video Joint-Embedding Predictive Architecture trained on 1M+ hours of videos to learn temporal and physical dynamics; for a $224 \times 224$ input, the encoder outputs $14 \times 14 \times 1024$ patch embeddings.
    
    \item \textbf{SigLIP~2~\cite{siglip2}:} An improved version of the Sigmoid Language-Image Pre-training model trained on the WebLI dataset (internet-scale image-text pairs); for a $224 \times 224$ input, the encoder outputs $14 \times 14 \times 768$ patch features.
\end{enumerate}

In this section, we present the comprehensive numerical results for benchmarking \method{} across the various pre-trained visual representations illustrated in \Cref{fig:vision}. For each configuration, we performed evaluations over three independent seeds, reporting the resulting mean and standard deviations. The performance for \method{}-VQ-BeT is detailed in \Cref{tab:vision-vqbet}, while the corresponding results for \method{}-Diffusion Policy are provided in \Cref{tab:vision-dp}.

\begin{table}[ht]
\centering
\caption{Effect of Pre-trained Visual Representations on~\method{} - VQ-BeT. Values within 2\% of the maximum performance for each task are shown in bold. }
\label{tab:vision-vqbet}
\begin{tabular}{lcccc}
\toprule
\textbf{Method} & \textbf{Push-T} & \textbf{LIBERO Goal} & \textbf{BlockPush} & \textbf{Cube} \\
\midrule
Ours -- DINOv2  & $\mathbf{0.69}$\,\std{0.01} & $\mathbf{0.96}$\,\std{0.01} & $1.20$\,\std{0.23}          & $1.35$\,\std{0.03}          \\
Ours -- DINOv3  & $0.65$\,\std{0.05}          & $\mathbf{0.95}$\,\std{0.02} & $0.96$\,\std{0.09}          & $0.96$\,\std{0.04}          \\
Ours -- WebSSL  & $\mathbf{0.68}$\,\std{0.02} & $\mathbf{0.94}$\,\std{0.01} & $\mathbf{1.68}$\,\std{0.15} & $\mathbf{1.68}$\,\std{0.02} \\
Ours -- V-JEPA2 & $0.65$\,\std{0.01}          & $0.86$\,\std{0.03}           & $1.46$\,\std{0.13}          & $1.36$\,\std{0.03}          \\
Ours -- SigLIP2 & $0.51$\,\std{0.01}          & $0.83$\,\std{0.01}          & $0.99$\,\std{0.10}          & $1.17$\,\std{0.03}          \\
\bottomrule
\end{tabular}
\end{table}

\begin{table}[ht]
\centering
\caption{Effect of Pre-trained Visual Representations on~\method{} - Diffusion Policy. Values within 2\% of the maximum performance for each task are shown in bold.}
\label{tab:vision-dp}
\begin{tabular}{lcccc}
\toprule
\textbf{Method} & \textbf{Push-T} & \textbf{LIBERO Goal} & \textbf{BlockPush} & \textbf{Cube} \\
\midrule
Ours -- DINOv2  & $\mathbf{0.81}$\,\std{0.01} & $\mathbf{0.98}$\,\std{0.00} & $1.25$\,\std{0.08}          & $1.24$\,\std{0.01}          \\
Ours -- DINOv3  & $0.73$\,\std{0.02}          & $0.94$\,\std{0.01}          & $1.22$\,\std{0.15}          & $1.17$\,\std{0.03}          \\
Ours -- WebSSL  & $\mathbf{0.80}$\,\std{0.01} & $\mathbf{0.98}$\,\std{0.00} & $\mathbf{1.65}$\,\std{0.08} & $\mathbf{1.73}$\,\std{0.02} \\
Ours -- V-JEPA2 & $0.72$\,\std{0.04}          & $0.91$\,\std{0.01}          & $1.60$\,\std{0.07} & $1.30$\,\std{0.05}          \\
Ours -- SigLIP2 & $0.64$\,\std{0.02}          & $0.83$\,\std{0.01}          & $1.43$\,\std{0.06}          & $1.23$\,\std{0.03}          \\
\bottomrule
\end{tabular}
\end{table}

\subsection{Hyperparameters}
\label{sec:misc}

We present the hyperparameters for \method{} training and relevant implementation details below. All hyperparameters of the \method{}-VQ-BeT variant and \method{}-Diffusion Policy variant have been reported in~\Cref{tab:vqbet_hyperparams} and~\Cref{tab:diffusion_hyperparams}. We further ablate \method{}'s model size in \cref{ssec:ablation_model_size}.

\begin{table}[ht]
\centering
\caption{\method{} - VQ-BeT Hyperparameters}
\label{tab:vqbet_hyperparams}
\resizebox{\textwidth}{!}{

\begin{tabular}{@{}lccccccc@{}}
  \toprule
  \textbf{Hyperparameter} & \textbf{Push-T} & \textbf{LIBERO Goal} & \textbf{BlockPush} &
  \textbf{Cube} & \textbf{Pickup} & \textbf{Open} & \textbf{Close} \\ \midrule
  \textit{VQ-VAE Configuration} & & & & & & & \\
  Latent Dimension    & 512          & 512          & 512      & 512          & 512          &
  512           & 512           \\
  Codebook Size       & 16            & 16           & 16      & 16           & 16           &
  16            & 16            \\
  Number of Groups          & 2            & 2            & 2      & 2            & 2
     & 2             & 2             \\ \midrule
  \textit{Architecture} & & & & & & & \\
  Number of Layers     ($N$)     & 8            & 6            & 8      & 8            & 8
          & 8             & 8             \\
  Number of Heads    ($n_{\text{heads}}$)       & 8            & 6            & 8       & 8
          & 8            & 8             & 8             \\
  Embedding Dimension  ($d_{\text{emb}}$)     & 512          & 120          & 512       & 512
        & 512          & 512           & 512                     \\ \midrule
  \textit{Policy} & & & & & & & \\
  Window Size               & 5            & 10           & 3      & 5            & 3
     & 3             & 3             \\
  Action Window             & 5            & 1            & 1      & 5            & 1
     & 1             & 1

  \\ \midrule
  \textit{Training} & & & & & & & \\
  Learning Rate             & \num{5e-5}   & \num{5e-5}   & \num{1e-4}      & \num{5e-5}   & \num{3e-4}
     & \num{3e-4}    & \num{3e-4}    \\
  Weight Decay              & \num{2e-4}   & \num{2e-4}   & 0      & \num{2e-4}   & \num{1e-4}
     & \num{1e-4}    & \num{1e-4}    \\
  Batch Size                & 8           & 32           & 32      & 128          & 256
     & 256           & 256

\\ \bottomrule

\end{tabular}
}
\end{table}

\begin{table}[ht]
\centering
\caption{\method{} - Real Robot VQ-BeT Hyperparameters}
\label{tab:vqbet_real_hyperparams}
\resizebox{0.7\textwidth}{!}{
\begin{tabular}{@{}lccc@{}}
  \toprule
  \textbf{Hyperparameter} & 
  \textbf{Cable Insertion} & 
  \textbf{Pen Collection} & 
  \textbf{Tool Hanging} \\ 
  \midrule

  \textit{VQ-VAE Configuration} & & & \\
  Latent Dimension & 512 & 512 & 512 \\
  Codebook Size & 16 & 16 & 16 \\
  Number of Groups & 2 & 2 & 2 \\
  \midrule

  \textit{Architecture} & & & \\
  Number of Layers ($N$) & 8 & 8 & 8 \\
  Number of Heads ($n_{\text{heads}}$) & 8 & 8 & 8 \\
  Embedding Dimension ($d_{\text{emb}}$) & 384 & 384 & 384 \\
  \midrule

  \textit{Policy} & & & \\
  Window Size & 2 & 2 & 2 \\
  Action Window & 10 & 5 & 5 \\
  \midrule

  \textit{Training} & & & \\
  Learning Rate & \num{3e-4} & \num{3e-4} & \num{3e-4} \\
  Weight Decay & \num{1e-4} & \num{1e-4} & \num{1e-4} \\
  Batch Size & 128 & 128 & 128 \\

  \bottomrule
\end{tabular}
}
\end{table}

\begin{table}[ht]
\centering
\caption{\method{} - Diffusion Policy Hyperparameters}
\label{tab:diffusion_hyperparams}
  \begin{tabular}{@{}lcccc@{}}
  \toprule
  \textbf{Hyperparameter} & \textbf{Push-T} & \textbf{LIBERO Goal} & \textbf{BlockPush} &
  \textbf{Cube} \\ \midrule
  \textit{Architecture} & & & & \\
  Number of Layers ($N$)  & 8            & 8            & 8      & 8            \\
  Number of Heads ($n_{\text{heads}}$)    & 4            & 4            & 8      & 4
    \\
  Embedding Dim  ($d_{\text{emb}}$)    & 256          & 256          & 512      & 256
  \\ \midrule
  \textit{Policy} & & & & \\
  Observation Horizon       & 2            & 2            & 3      & 2            \\
  Action Horizon            & 5            & 3            & 1      & 3            \\
  \midrule
  \textit{Training} & & & & \\
  Learning Rate             & \num{1e-4}   & \num{5e-5}   & \num{1e-4}      & \num{1e-4}   \\
  Weight Decay              & 0            & \num{2e-4}   & 0      & 0            \\
  Batch Size                & 256          & 256          & 256      & 256          \\
  \bottomrule
\end{tabular}
\end{table}

\subsection{Additional ablations}
\label{sec:ablations}
\subsubsection{\method{} Attention Mask Ablation}
\label{sec:attention_ablation}

Transformer-based policies traditionally condition on a single token per observation step, allowing direct application of standard token-level causal attention. In \method{}, each observation is now a sequence of tokens and we thus apply a block-causal attention mask as mentioned in \Cref{sec:policy}. Under this setting, every patch token attends fully to all other patch tokens within the same frame, while attention across frames remains strictly causal. In this section, we ablate this block-causal attention along with two other masking choices: 1) full attention, with no masking; 2) token-causal, the naive implementation that treats every patch as an independent timestep. 

As shown in~\Cref{tab:attention_ablation}, block-causal matches or exceeds the alternatives on most (policy head, task) pair, with the largest gains on Diffusion Policy. Full attention lets every position attend to patches from future frames during training, thus violating the structure of the underlying task, mildly degrading performance. Token-causal preserves causality but slices the conditioning within a frame. The downstream effect depends on where in the sequence the policy's loss is applied. For VQ-BeT, predictions are read out from the last patch of each frame, so under token-causal it already attends to the same context as block-causal. Therefore, the two are nearly tied. For Diffusion Policy, the denoising loss is applied at every decoder position: early positions see only a partial fraction of the first frame's patches, causing significant performance drop on Cube. 

\begin{table}[ht]
\centering
\caption{Attention mask ablation for \method{}. Block-causal matches or outperforms alternative masking choices across two policy heads on two environments.}
\label{tab:attention_ablation}
\begin{tabular}{lcccc}
\toprule
 & \multicolumn{2}{c}{Ours - VQ-BeT} & \multicolumn{2}{c}{Ours - DP} \\
\cmidrule(lr){2-3}\cmidrule(lr){4-5}
Mask & Push-T & Cube & Push-T & Cube \\
\midrule
Full                & 0.64          & 1.09          & \textbf{0.83} & 1.10          \\
Token-causal        & \textbf{0.73} & 1.36          & 0.70          & 0.11          \\
Block-causal (ours) & 0.70          & \textbf{1.38} & \textbf{0.83} & \textbf{1.24} \\
\bottomrule
\end{tabular}
\end{table}

\subsubsection{\method{} Model Size Ablation}
\label{ssec:ablation_model_size}

In this section, we investigate how \method{}'s performance scales with model size for both the VQ-BeT and Diffusion Policy variants. We sweep for different number of layers ($N$), number of heads ($n_{\text{heads}}$), and embedding dimension ($d_{\text{emb}}$) of the transformer policy backbone. As detailed in \Cref{tab:model-size-ablation}, performance consistently scales with increased model capacity across both policy variants. 

\begin{table}[ht]
\centering
\caption{Model Size Ablation: Architectural Parameters and Performance on Push-T for \method{} with DINOv2 (ViT-S) patch features. We report the final mean coverage of the T-shaped block over the target T across 100 evaluation rollouts.}
\label{tab:model-size-ablation}
\begin{tabular}{lccccc}
\toprule
\textbf{Method} & $N$ & $n_{\text{heads}}$ & $d_{\text{emb}}$ & Size & Final Coverage ($\uparrow$) \\
\midrule
Ours -- VQ-BeT           & 4 & 4 & 64  & 25.62M & 0.50 \\
Ours -- VQ-BeT           & 6 & 6 & 120 & 26.64M & 0.57 \\
Ours -- VQ-BeT           & 8 & 8 & 512 & 51.55M & 0.69 \\
\midrule
Ours -- Diffusion Policy & 4 & 4 & 64  & 22.77M & 0.07 \\
Ours -- Diffusion Policy & 6 & 6 & 120 & 25.30M & 0.56 \\
Ours -- Diffusion Policy & 8 & 4 & 256 & 40.43M & 0.83 \\
\bottomrule
\end{tabular}
\end{table}

\subsection{Additional figures and details}

\begin{figure*}[t]
    \centering
    \includegraphics[width=\textwidth]{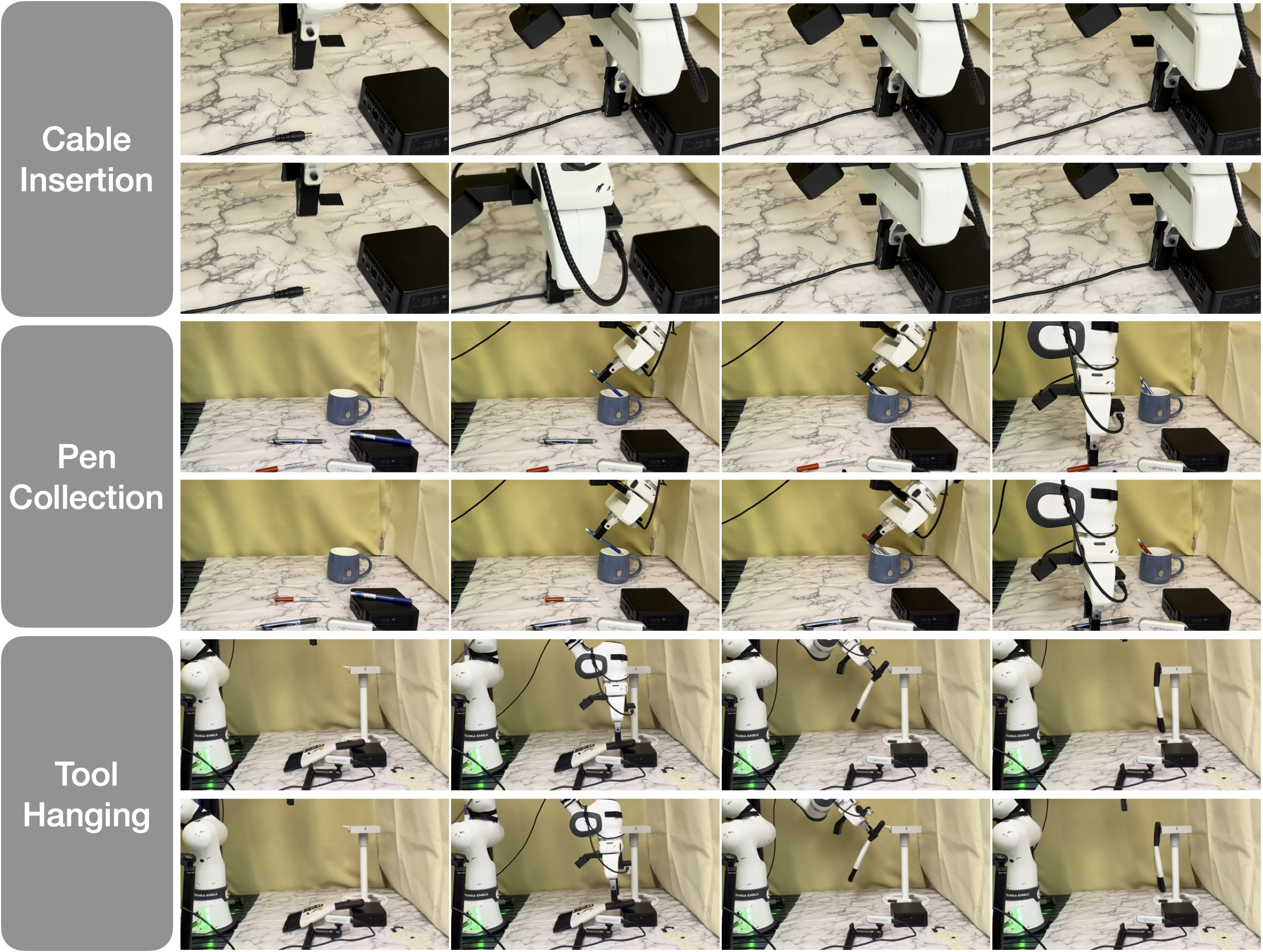}
    \caption{Successful rollouts of \method{} for Cable Insertion, Pen Collection, and Tool Hanging.}
    \label{fig:real-success}
\end{figure*}

\begin{figure*}[t]
    \centering
    \includegraphics[width=\textwidth]{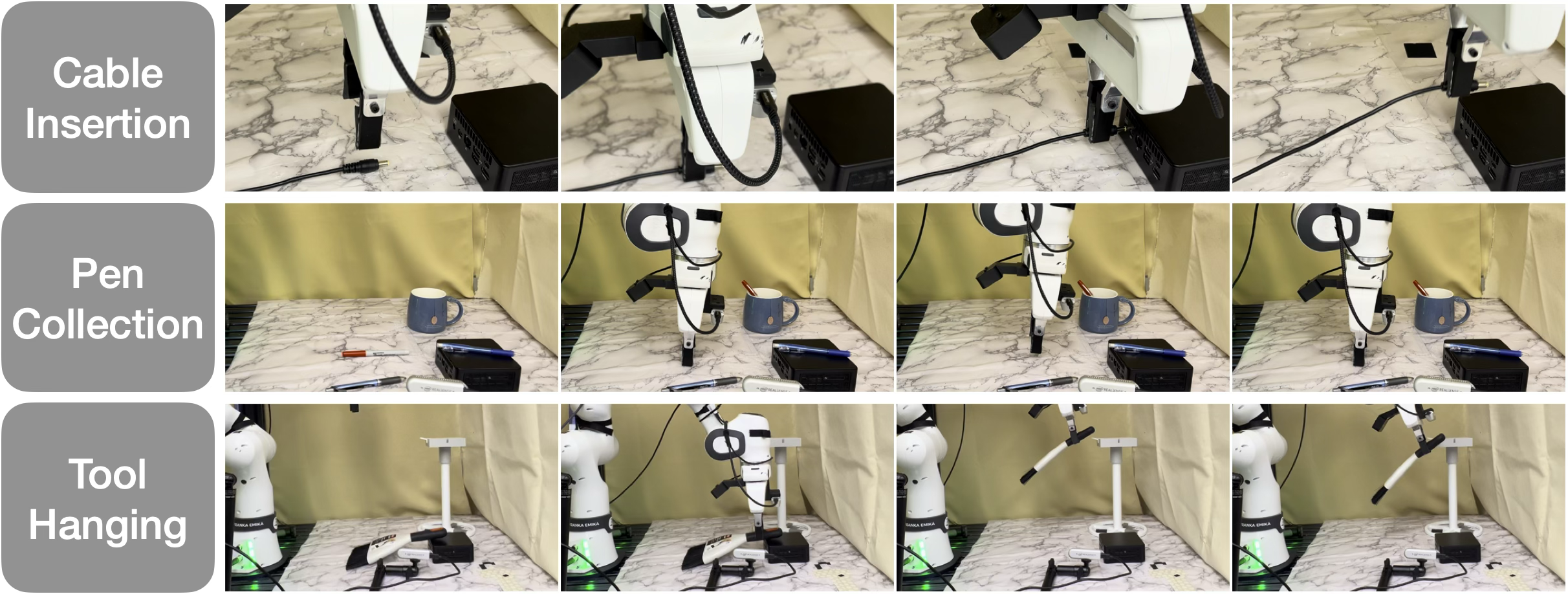}
    \caption{Failure rollouts of \method{} for Cable Insertion, Pen Collection, and Tool Hanging.}
    \label{fig:real-failure}
\end{figure*}

\Cref{fig:real-success} shows representative successful rollout trajectories of \method{}-VQ-BeT for the three real-world manipulation tasks: Cable Insertion, Pen Collection, and Tool Hanging. \Cref{fig:real-failure} shows representative failure rollouts for the same tasks.

\begin{figure}[t]
\center
\includegraphics[width=0.9\textwidth]{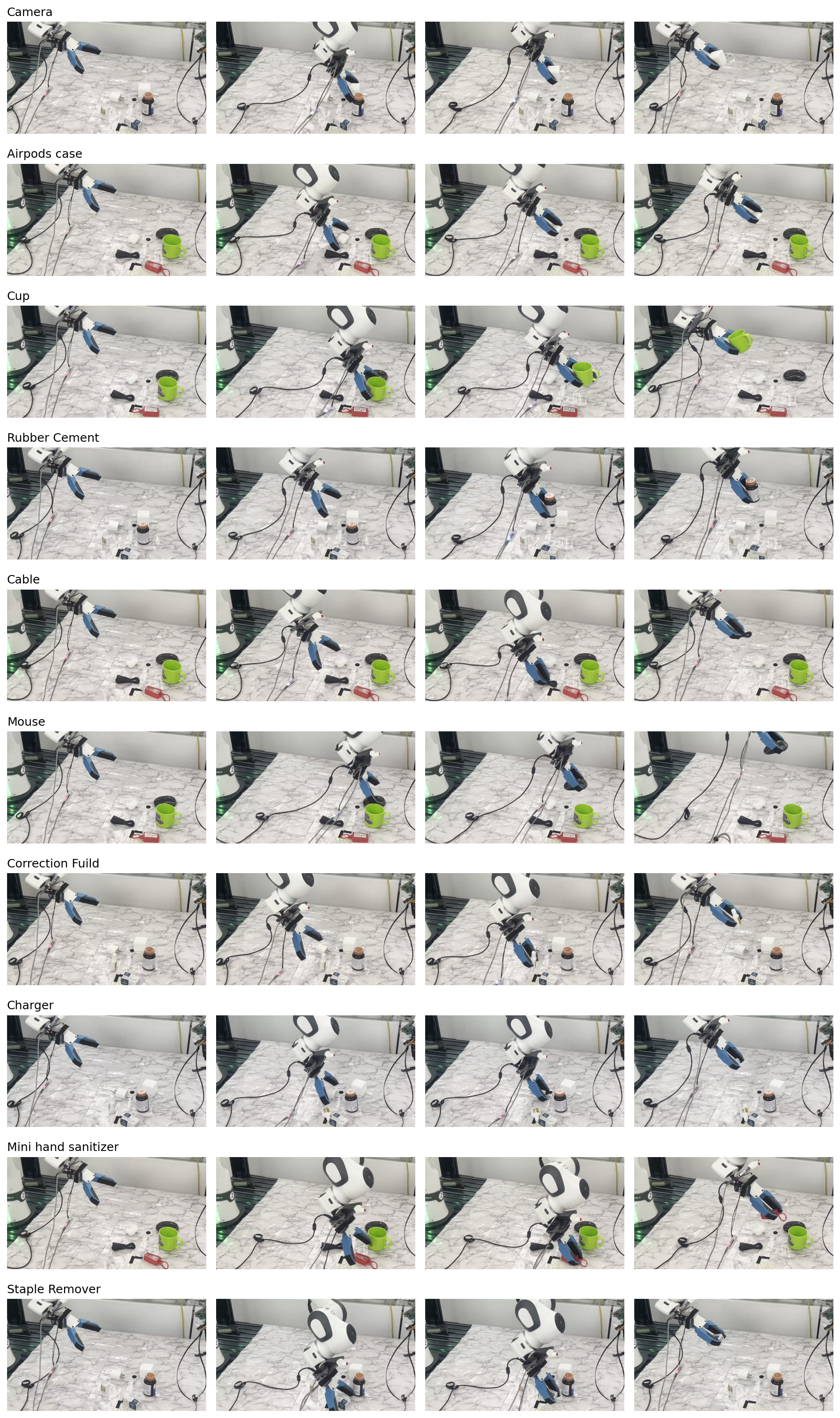}
\caption{CAP real Franka object pickup successes.}
\label{fig:cap-success}
\end{figure}

\begin{figure*}[t]
\center
\includegraphics[width=\textwidth]{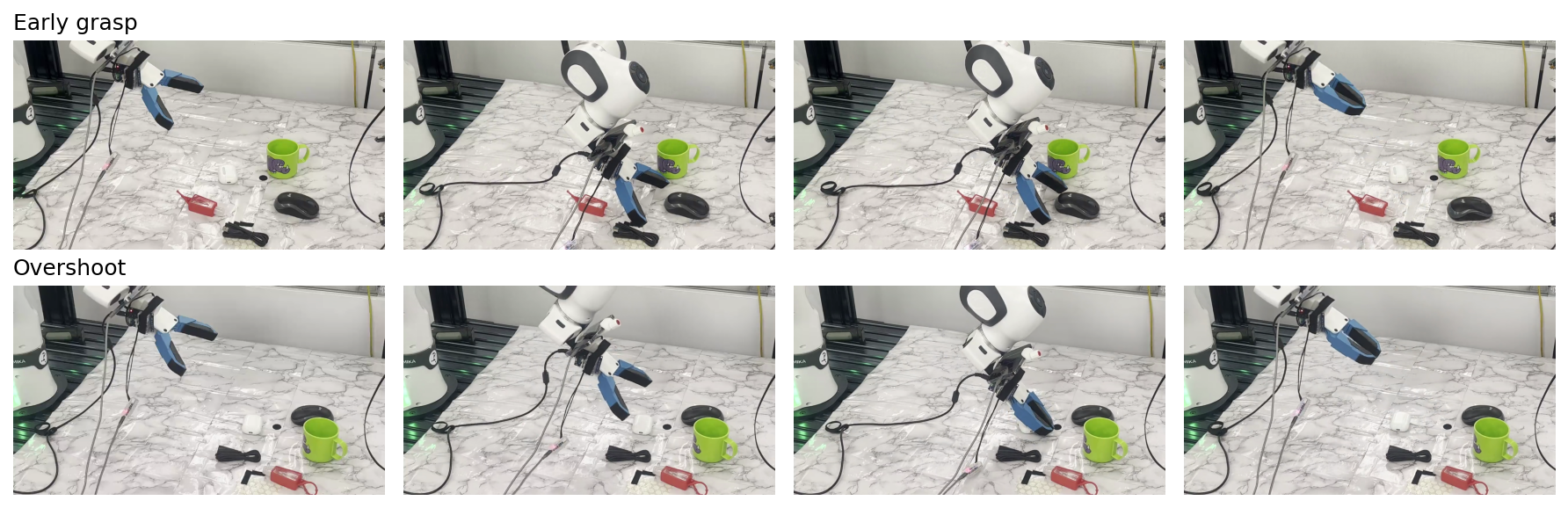}
\caption{CAP real Franka object pickup failure modes.}
\label{fig:cap-failure}
\end{figure*}

\Cref{fig:cap-success} shows rollout success trajectories for each object in the Real Franka object pickup experiment. \Cref{fig:cap-failure} shows rollout failure modes (early grasping, and overshooting) in the Real Franka object pickup experiment.

\begin{figure*}[t]
\center
\includegraphics[width=0.9\textwidth]{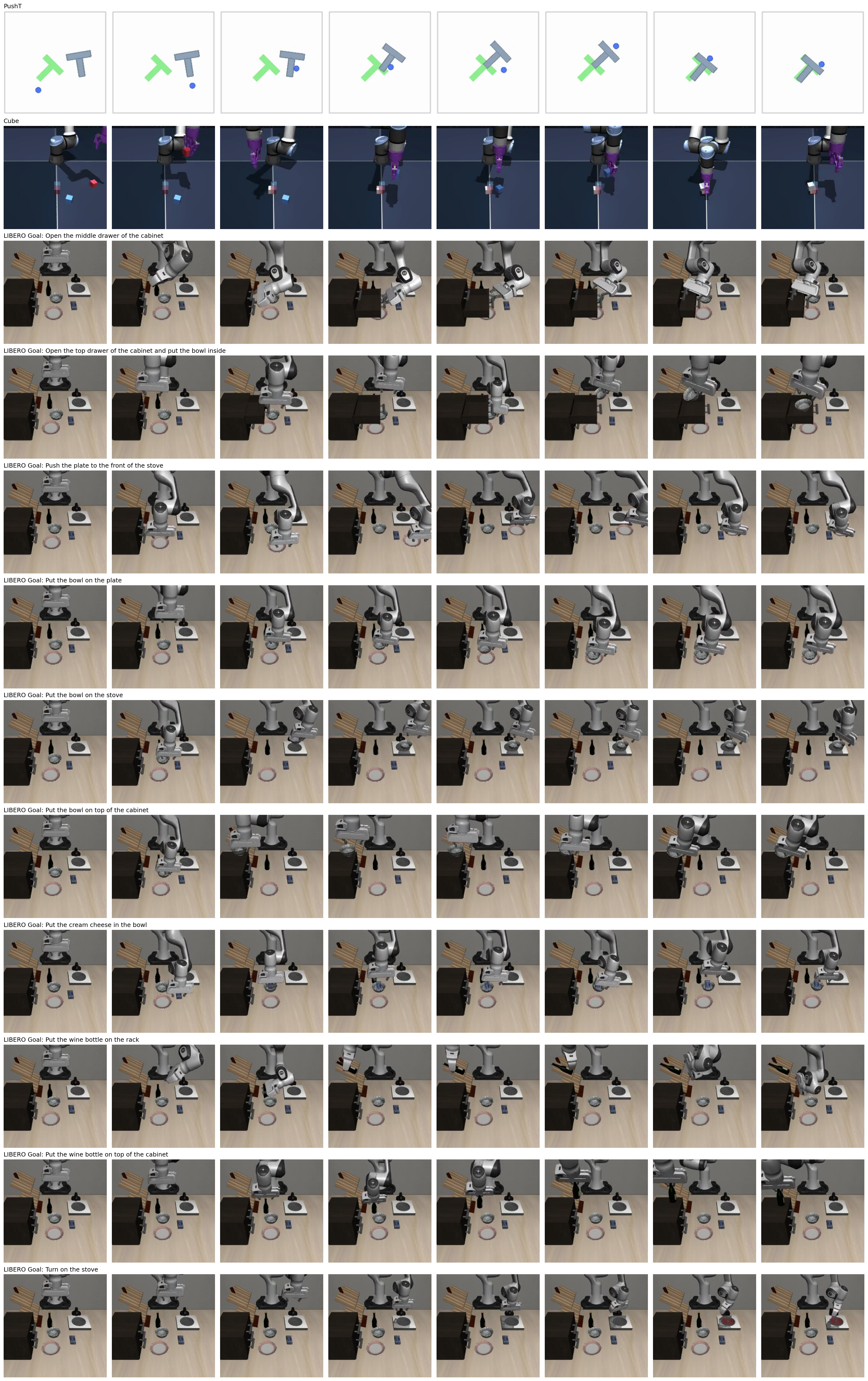}
\caption{Push-T, Cube, and LIBERO Goal environment Ours - VQ-BeT evaluation rollouts.}
\label{fig:sim-evals}
\end{figure*}

\Cref{fig:sim-evals} shows \method{}-VQ-BeT rollout trajectories for Push-T, Cube, and LIBERO Goal.

\begin{figure*}[t]
\center
\includegraphics[width=\textwidth]{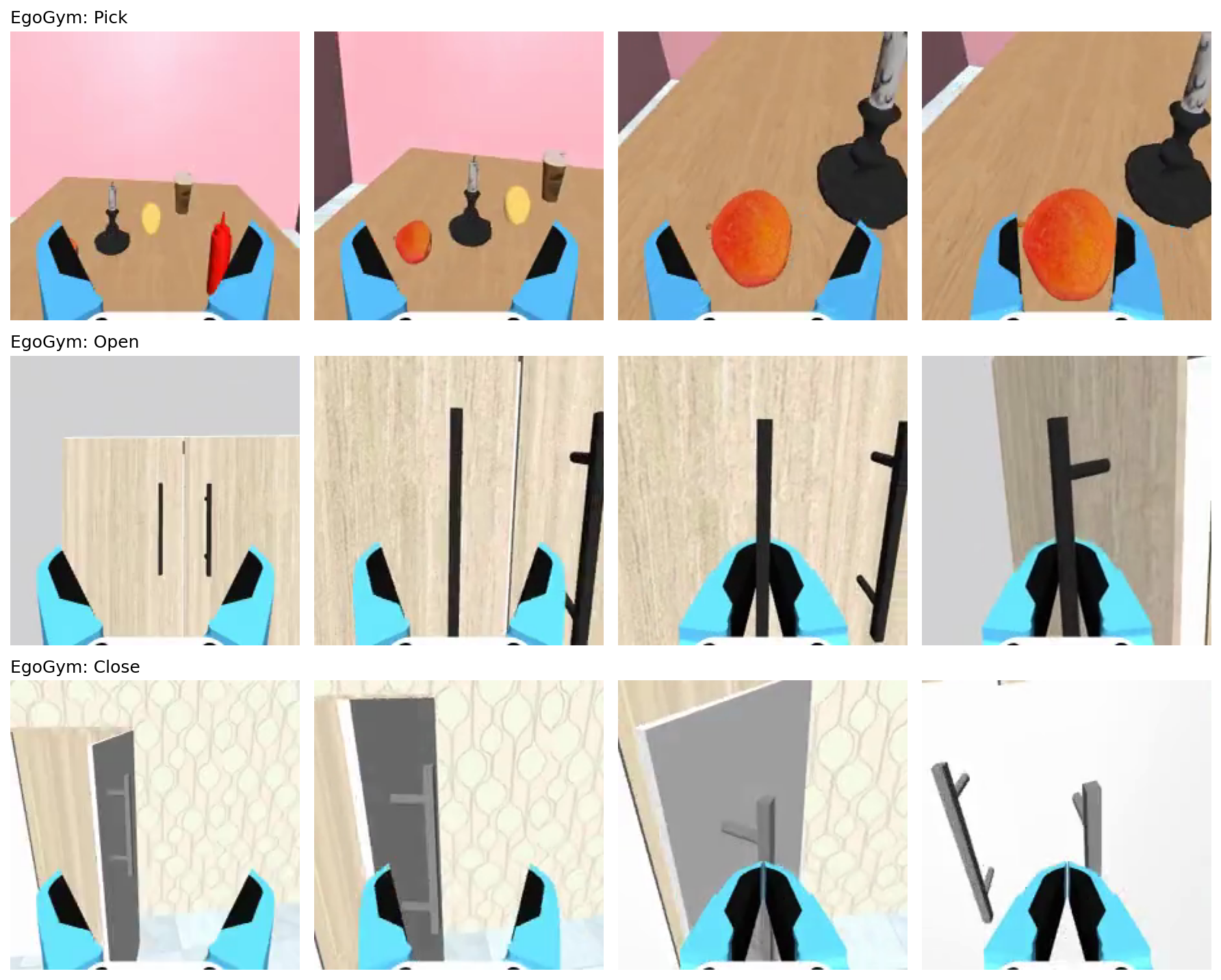}
\caption{EgoGym pick, open, and close \method{} evaluation rollouts.}
\label{fig:egogym-evals}
\end{figure*}

\Cref{fig:egogym-evals} shows \method{}-VQ-BeT rollout trajectories for object pickup, door opening, and closing tasks in EgoGym.

\begin{figure*}[t]
\center
\includegraphics[width=\textwidth]{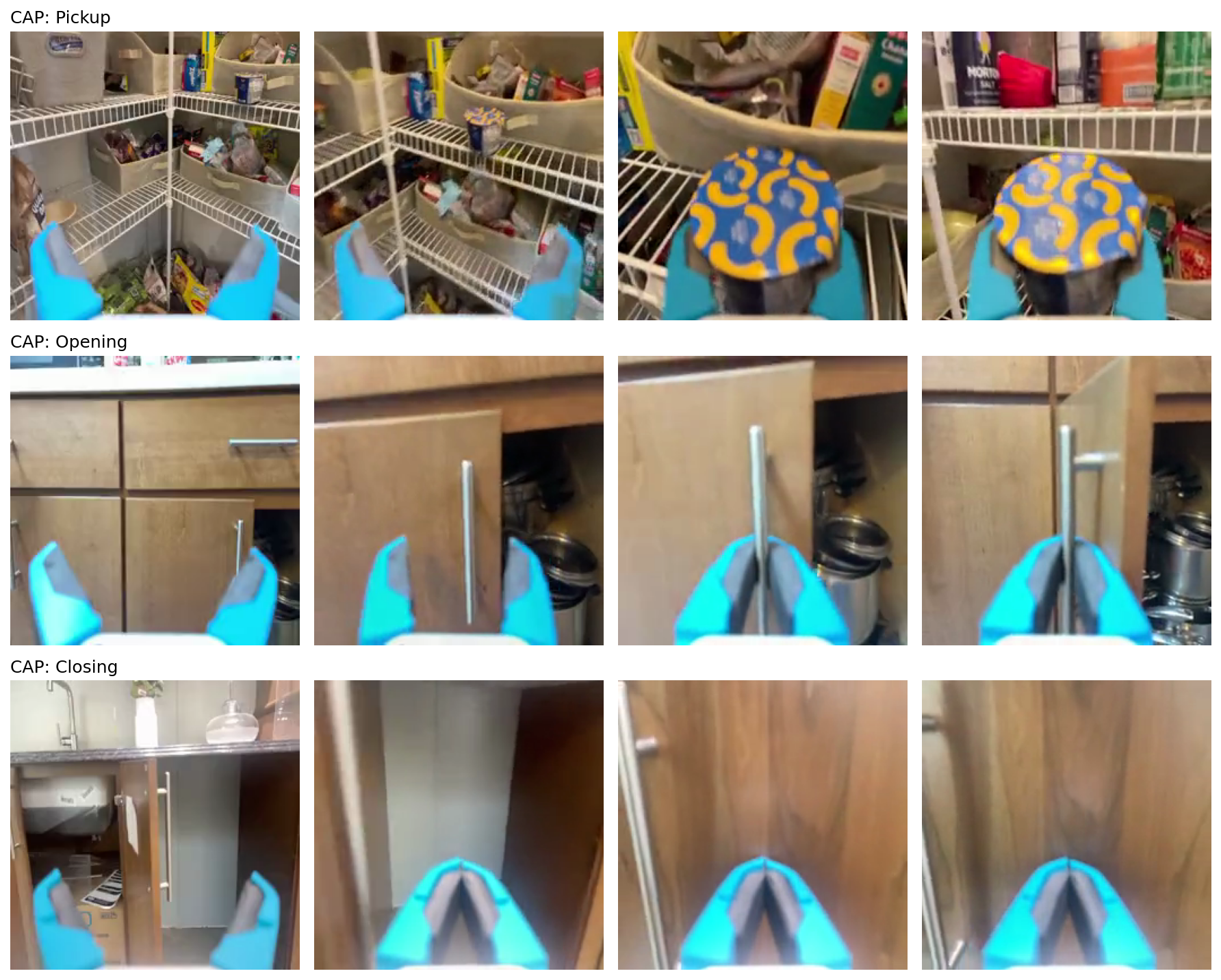}
\caption{CAP dataset pick, open, and close trajectory samples.}
\label{fig:cap-dataset-samples}
\end{figure*}

\Cref{fig:cap-dataset-samples} shows sample trajectories of object pickup, opening, and closing from the CAP dataset.

\end{document}